\documentclass[11pt]{article}

\usepackage{epsfig}
\usepackage{a4}
\usepackage{nobi}
\usepackage{ims}
\usepackage{nobiDG}
\usepackage{colacl98}
\usepackage{nobibib}

\newcounter{defcnt}
\newenvironment{definition}[2][]
        {\refstepcounter{defcnt}
                \par
                \noindent
                {\bf Def:} 
                \itshape%
	}
        {\par
	}

\begin{document}

\title{\appearedIn{In: Proc. of COLING-ACL'98, pages 174--180}{174}
	Separating Surface Order and Syntactic Relations \\
	in a Dependency Grammar}
\author{Norbert Br\"{o}ker \\ \nobijoborga \\ \nobijobstrasse \\
	\nobijobplz\ \nobijobstadt \\ \textsc{\nobijobemail} }





\maketitle

\begin{abstract}
This paper proposes decoupling the dependency tree from word order, such
that surface ordering is not determined by traversing the dependency tree.
We develop the notion of a \emph{word order domain structure}, which is
linked but structurally dissimilar to the syntactic dependency tree. The
proposal results in a lexicalized, declarative, and formally precise
description of word order; features which lack previous proposals for
dependency grammars.  Contrary to other lexicalized approaches to word
order, our proposal does not require lexical ambiguities for ordering
alternatives. 
\end{abstract}

\section{Introduction
	\label{ch:intro}}
Recently, the concept of valency has gained considerable attention. Not
only do all linguistic theories refer to some reformulation of the
traditional notion of valency (in the form of $\theta$-grid,
subcategorization list, argument list, or extended domain of locality);
there is a growing number of parsers based on binary relations between
words \cite{Eisner1997,Maruyama1990}. 

Given this interest in the valency concept, and the fact that word order is
one of the main difference between phrase-structure based approaches
(henceforth PSG) and dependency grammar (DG), it is valid to ask whether DG
can capture word order phenomena without recourse to phrasal nodes, traces,
slashed categories, etc. A very early result on the weak generative
equivalence of context-free grammars and DGs suggested that DGs are
incapable of describing surface word order \cite{Gaifman1965}. This result
has recently been critizised to apply only to impoverished DGs which do not
properly represent formally the expressivity of contemporary DG variants
\cite{nobi:acl97}.

Our position will be that dependency relations are motivated semantically
\cite{Tesniere1959}, and need not be projective (i.e., may cross if
projected onto the surface ordering). We argue for so-called
\emph{word order domains}, consisting of partially ordered sets of words
and associated with nodes in the dependency tree. These order domains
constitute a tree defined by set inclusion, and surface word order is
determined by traversing this tree. A syntactic analysis therefor consists
of two linked, but dissimilar trees.

Sec.~\ref{ch:dg} will briefly review approaches to word order in DG. In
Sec.~\ref{ch:domains}, word order domains will be defined, and
Sec.~\ref{ch:logic} introduces a modal logic to describe dependency
structures. Sec.~\ref{ch:example} applies our approach to the German clause
and Sec.~\ref{ch:compare} relates it to some PSG approaches.

\section{Word Order in DG
	\label{ch:dg}}

A very brief characterization of DG is that it recognizes only lexical, not
phrasal nodes, which are linked by directed, typed, binary relations to
form a dependency tree \cite{Tesniere1959,Hudson1993}. 
The following overview of DG flavors shows that various mechanisms (global
rules, general graphs, procedural means) are generally employed to lift the
limitation of projectivity and discusses some shortcomings of these
proposals.

\paragraph{Functional Generative Description}
\cite{Sgall1986} assumes a language-independent \emph{underlying order},
which is represented as a projective dependency tree. This abstract
representation of the sentence is mapped via \emph{ordering rules} to the
concrete surface realization. Recently, \textcite{Kruijff1997} has given a
categorial-style formulation of these ordering rules. He assumes
associative categorial operators, permuting the arguments to yield the
surface ordering. One difference to our proposal is that we argue for a
representational account of word order (based on valid structures
representing word order), eschewing the non-determinism introduced by unary
operators; the second difference is the avoidance of an underlying
structure, which stratifies the theory and makes incremental processing
difficult.

\paragraph{Meaning-Text Theory}
\cite{Melcuk1988} assumes seven \emph{strata of representation}. The rules
mapping from the unordered dependency trees of surface-syntactic
representations onto the annotated lexeme sequences of deep-morphological
representations include \emph{global ordering rules} which allow
discontinuities. These rules have not yet been formally specified
\cite[p.187f]{Melcuk+Pertsov1987}.

\paragraph{Word Grammar}
(WG, \textcite{Hudson1990}) is based on \emph{general graphs} instead of
trees. The ordering of two linked words is specified together with their
dependency relation, as in the proposition ``\texttt{object of verb follows
it}''. Extraction of, e.g., objects is analyzed by establishing an
additional dependency called \texttt{visitor} between the verb and the
extractee, which requires the reverse order, as in ``\texttt{visitor of
verb precedes it}''. This results in inconsistencies, since an extracted
object must follow the verb (being its \texttt{object}) and at the same
time precede it (being its \texttt{visitor}). 
The approach compromises the semantic motivation of dependencies by adding
\emph{purely order-induced dependencies}. WG is similar to our proposal in
that it also distinguishes a propositional meta language describing the
graph-based analysis structures.

\paragraph{Dependency Unification Grammar}
(DUG, \textcite{Hellwig1986}) defines a tree-like data structure for the
representation of syntactic analyses. Using morphosyntactic features with
special interpretations, a word defines \emph{abstract positions} into
which modifiers are mapped. Partial orderings and even discontinuities can
thus be described by allowing a modifier to occupy a position defined by
some transitive head. The approach requires that the \emph{parser}
interpretes several features specially, and it cannot restrict the scope of
discontinuities.

\paragraph{Slot Grammar}
\cite{McCord1990} employs a number of rule types, some of which are
exclusively concerned with precedence. So-called head/slot and slot/slot
\emph{ordering rules} describe the precedence in projective trees,
referring to arbitrary predicates over head and modifiers. Extractions
(i.e., discontinuities) are merely handled by a mechanism built into the
\emph{parser}.

\section{Word Order Domains
	\label{ch:domains}}

Summarizing the previous discussion, we require the following of a word
order description for DG:

\begin{itemize}
\item not to compromise the semantic motivation of dependencies,
\item to be able to restrict discontinuities to certain constructions and
delimit their scope,
\item to be lexicalized without requiring lexical ambiguities for the
representation of ordering alternatives,
\item to be declarative (i.e., independent of an analysis procedure), and
\item to be formally precise and consistent.
\end{itemize}

The subsequent definition of an order domain structure and its linking to
the dependency tree satisify these requirements.

\subsection{The Order Domain Structure}

A \emph{word order domain} is a set of words, generalizing the notion of
positions in DUG. The cardinality of an order domain may be restricted to
at most one element, at least one element, or -- by conjunction -- to
exactly one element. Each word is associated with a sequence of order
domains, one of which must contain the word itself, and each of these
domains may require that its elements have certain features. Order domains
can be partially ordered based on set inclusion: If an order domain $d$
contains word $w$ (which is not associated with $d$), every word $w'$
contained in a domain $d'$ associated with $w$ is also contained in $d$;
therefor, $d' \subset d$ for each $d'$ associated with $w$. This partial
ordering induces a tree on order domains, which we call the \emph{order
domain structure}.

\begin{figure}[t]
\centering
\epsfig{file=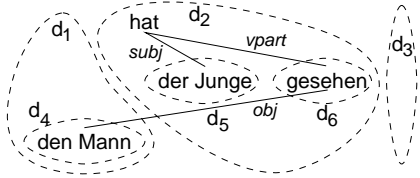,width=0.7\columnwidth}
\caption{Dependency Tree and Order Domains for \os{Den Mann hat der Junge
	gesehen} \label{fig:dep-tree}}
\end{figure}

\begin{figure}[t]
\centering
\epsfig{file=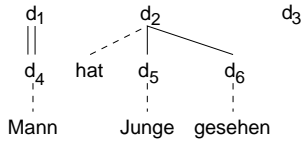,width=0.5\columnwidth}
\caption{Order Domain Structure for \os{Den Mann hat der Junge gesehen}
	\label{fig:dom-tree}}
\end{figure}

Take the example of German \os{Den Mann hat der Junge gesehen} (\os{the
man$_{ACC}$ -- has -- the boy$_{NOM}$ -- seen}). Its dependency tree is
shown in Fig.\ref{fig:dep-tree}, with word order domains indicated by
dashed circles. The finite verb, \os{hat}, defines a sequence of domains,
\tuple{d_1, d_2, d_3}, which roughly correspond to the topological fields
in the German main clause. The nouns \os{Mann} and \os{Junge} and the
participle \os{gesehen} each define one order domain ($d_4, d_5, d_6$,
resp.). Set inclusion gives rise to the domain structure in
Fig.\ref{fig:dom-tree}, where the individual words are attached by dashed
lines to their including domains ($d_1$ and $d_4$ collapse, being
identical).%
\footnote{ 
Note that in this case, we have not a single rooted tree, but rather an
ordered sequence of trees (by virtue of ordering $d_1, d_2$, and $d_3$) as
domain structure. In general, we assume the sentence period to govern the
finite verb and to introduce a single domain for the complete sentence. 
}

\subsection{Surface Ordering}

How is the surface order derived from an order domain structure? First of
all, the ordering of domains is inherited by their respective elements,
i.e., \os{Mann} precedes (any element of) $d_2$, \os{hat} follows (any
element of) $d_1$, etc.

Ordering within a domain, e.g., of \os{hat} and $d_6$, or $d_5$ and $d_6$,
is based on precedence predicates (adapting the precedence predicates of
WG). There are two different types, one ordering a word w.r.t. any other
element of the domain it is associated with (e.g., \os{hat} w.r.t. $d_6$),
and another ordering two modifiers, referring to the dependency relations
they occupy ($d_5$ and $d_6$, referring to \texttt{subj} and
\texttt{vpart}). A verb like \os{hat} introduces two precedence predicates,
requiring other words to follow itself and the participle to follow subject
and object, resp.:%
\footnote{
For details of the notation, please refer to Sec.~\ref{ch:logic}.
}

\begin{center}
\os{hat} $\IMPL (<_{*} \AND\ \tuple{\texttt{vpart}} >_{\{\texttt{subj}, \texttt{obj}\}})$
\end{center}

Informally, the first conjunct is satisfied by any domain in which no word
precedes \os{hat}, and the second conjunct is satisfied by any domain in
which no subject or object follows a participle. The domain structure in
Fig.\ref{fig:dom-tree} satisfies these restrictions since nothing follows
the participle, and because \os{den Mann} is not an element of $d_2$, which
contains \os{hat}. This is an important interaction of order domains and
precedence predicates: Order domains define scopes for precedence
predicates. In this way, we take into account that dependency trees are
flatter than PS-based ones%
\footnote{
Note that each phrasal level in PS-based trees defines a scope for linear
precedence rules, which only apply to sister nodes. 
}
and avoid the formal inconsistencies noted above for WG.

\subsection{Linking Domain Structure and Dependency Tree}

Order domains easily extend to discontinuous dependencies. Consider the
non-projective tree in Fig.\ref{fig:dep-tree}. Assuming that the finite
verb governs the participle, no projective dependency between the object
\os{den Mann} and the participle \os{gesehen} can be established. We allow
non-projectivity by loosening the linking between dependency tree and
domain structure: A modifier (e.g., \os{Mann}) may not only be inserted
into a domain associated with its direct head (\os{gesehen}), but also into
a domain of a transitive head (\os{hat}), which we will call the
\emph{positional head}.

The possibility of inserting a word into a domain of some transitive head
raises the questions of how to require contiguity (as needed in most
cases), and how to limit the distance between the governor and the modifier
in the case of discontinuity. From a descriptive viewpoint, the
\emph{syntactic construction} is often cited to determine the possibility
and scope of discontinuities \cite{Bhatt1990,Matthews1981}. In PS-based
accounts, the construction is represented by phrasal categories, and
extraction is limited by bounding nodes (e.g.,
\textcite{Haegeman1994,Becker1991}). In dependency-based accounts, the
construction is represented by the dependency relation, which is typed or
labelled to indicate constructional distinctions which are
configurationally defined in PSG. Given this correspondence, it is natural
to employ dependencies in the description of discontinuities as follows:
For each modifier of a certain head, a set of dependency types is defined
which may link the direct head and the positional head of the modifier
(\os{gesehen} and \os{hat}, resp.). If this set is empty, both heads are
identical and a contiguous attachment results. The impossibility of
extraction from, e.g., a finite verb phrase may follow from the fact that
the dependency embedding finite verbs, \texttt{propo}, may not appear on
any path between a direct and a positional head.%
\footnote{%
One review pointed out that some verbs may allow extractions, i.e., that
this restriction is lexical, not universal. This fact can easily be
accomodated because the possibility of discontinuity (and the dependency
types across which the modifier may be extracted) is described in the
lexical entry of the verb. In fact, a universal restriction could not even
be stated because the treatment is completely lexicalized. 
}

\section{The Description Language
	\label{ch:logic}}

This section sketches a logical language describing the dependency
structure. It is based on modal logic and owes much to work of
\textcite{Blackburn-SLT}. As he argues, standard Kripke models can be
regarded as directed graphs with node annotations. We will use this
interpretation to represent dependency structures. Dependencies and the
mapping from dependency tree to order domain structure are described by
modal operators, while simple properties such as word class, features, and
cardinality of order domains are described by modal propositions.

\subsection{Model Structures}

In the following, we assume a set of words, \W, ordered by a precedence
relation, \pre, a set of dependency types, \D, a set of atomic feature
values \A, and a set of word classes, \C. We define a family of dependency
relations $R_d \subset \W \times \W, d \in \D$ and for convenience
abbreviate the union $\bigcup_{d \in \D}R_d$ as $R_\D$.

\begin{definition}{Dependency Tree}
A \emph{dependency tree} is a tuple \tuple{\W, w_r, R_\D, V_\A, V_\C},
where $R_\D$ forms a tree over \W\ rooted in $w_r$, $V_\A: \W \mapsto 2^\A$
maps words to sets of features, and $V_\C: \W \mapsto \C$ maps words to
word classes.
\end{definition}

\begin{definition}{Order Domain}
An \emph{order domain (over \W)} $m$ is a set of words from \W\ where
$\forall w_1, w_2, w_3 \in \W: (w_1 \pre w_2 \pre w_3 \AND w_1 \in m \AND
w_3 \in m) \IMPL w_2 \in m$.
\end{definition}

\begin{definition}{Order Domain Structure}
An \emph{order domain structure (over \W)} \M\ is a set of order domains
where $\forall m, m' \in \M: m \cap m' = \emptyset \OR m \subseteq m' \OR
m' \subseteq m$.
\end{definition}

\begin{definition}{Dependency Structure}
A \emph{dependency structure} $T$ is a tuple \tuple{\W, w_r, R_\D, V_\A,
V_\C, \M, V_\M} where \tuple{\W, w_r, R_\D, V_\A, V_\C} is a dependency
tree, \M\ is an order domain structure over \W, and $V_\M: \W \mapsto \M^n$
maps words to order domain sequences.
\end{definition}

Additionally, we require for a dependency structure four more conditions:
(1) Each word $w$ is contained in exactly one of the domains from
$V_\M(w)$, (2) all domains in $V_\M(w)$ are pairwise disjoint, (3) each
word (except $w_r$) is contained in at least two domains, one of which is
associated with a (transitive) head, and (4) the (partial) ordering of
domains (as described by $V_\M$) is consistent with the precedence of the
words contained in the domains (see \cite{nobi:diss} for more details).

\subsection{The Language \Ld}

\begin{figure*}[th]
\centering \small
$
\begin{array}{|r@{\ \in \Ld,\ }l|l@{\ T,w \SAT\ }l@{\ :\Leftrightarrow\ }l|}
\hline
\multicolumn{2}{|c|}{\mbox{Syntax (valid formulae)}} 
	& \multicolumn{3}{c|}{\mbox{Semantics (satisfaction relation)}} \\
\hline
c & \forall c \in \C & & c & c = V_\C(w) \\
a & \forall a \in \A & & a & a \in V_\A(w) \\
\tuple{d} \phi & \forall d \in \D, \phi \in \Ld & & \tuple{d} \phi 
	& \exists w' \in \W: wR_dw' \AND T,w' \SAT \phi \\
<_{*} & & & <_{*} & \begin{array}[t]{l}
			\exists m \in \M: (V_\M(w) = \tuple{\cdots m \cdots} \\
			\AND \forall w' \in m: (w = w' \OR w \pre w'))
		    \end{array} \\
<_\delta & \forall \delta \subseteq \D & & <_\delta & \begin{array}[t]{l}
						\neg \exists w', w'', w''' \in \W: places(w', w) \\
						\AND places(w', w'') \AND w''' R_\delta w \AND w''' \pre w
					      \end{array} \\
\disko_\delta & \forall \delta \subset \D & & \disko_\delta & \begin{array}[t]{l}
								\exists w', w'' \in \W: w R_\D w \AND \\
								places(w'', w) \AND w'' R_\delta^* w'o
							      \end{array} \\
\modM{i} \texttt{single} & \forall i \in \N & & \modM{i} \texttt{single}
	& \size{ \left\{ w' \left| 
			\begin{array}{l}
			w' \in \proj{i}(V_\M(w)) \AND \\
			\neg\exists w'': (w'' R_\D w' \AND \\ 
			w'' \in \proj{i}(V_\M(w)))
			\end{array} \right. \hspace{-.8em} \right\} } \leq 1 \\
\modM{i} \texttt{filled} & \forall i \in \N & & \modM{i} \texttt{filled}
	& \mid\proj{i}(V_\M(w))\mid \geq 1 \\
\modallM{i} a & \forall i \in \N, a \in \A & & \modallM{i} a 
	& \forall w' \in \proj{i}(V_\M(w)): T,w' \SAT a \\
\phi \AND \psi & \forall \phi, \psi \in \Ld & & \phi \AND \psi
	& T,w \SAT \phi \mbox{ and } T,w \SAT \psi \\
\neg \phi & \forall \phi \in \Ld & & \neg \phi & \mbox{ not } T,w \SAT \psi \\
\hline
\end{array}
$
\caption{Syntax and Semantics of \Ld\ Formulae
	\label{fig:Ld}}
\end{figure*}

Fig.\ref{fig:Ld} defines the logical language \Ld\ used to describe
dependency structures. Although they have been presented differently, they
can easily be rewritten as (multimodal) Kripke models: The dependency
relation $R_d$ is represented as modality $\tuple{d}$ and the mapping from
a word to its $i$th order domain as modality $\modM{i}$.%
\footnote{
The modality $\modallM{i}$ can be viewed as an abbreviation of
$\modM{i}\modallM{}$, composed of a mapping from a word to its $i$th order
domain and from that domain to all its elements.
}
All other formulae denote properties of nodes, and can be formulated as
unary predicates -- most evident for word class and feature assignment. For
the precedence predicates $<_{*}$ and $<_\delta$, there are inverses
$>_{*}$ and $>_\delta$. For presentation, the relation $places \subset \W
\times \W$ has been introduced, which holds between two words iff the first
argument is the positional head of the second argument.

\label{ch:levels}
A more elaborate definition of dependency structures and \Ld\ defines two
more dimensions, a feature graph mapped off the dependency tree much like
the proposal of \textcite{Blackburn-SLT}, and a conceptual representation
based on terminological logic, linking content words with reference objects
and dependencies with conceptual roles.

\section{The German Clause
	\label{ch:example}}

Traditionally, the German main clause is described using three topological
fields; the initial and middle fields are separated by the finite
(auxiliary) verb, and the middle and the final fields by infinite verb
parts such as separable prefixes or participles. We will generalize this
field structure to verb-initial and verb-final clauses as well, without
going into the linguistic motivation due to space limits. 

\begin{figure}[t]
\centering \small
$
\begin{array}{@{}r@{\,}ll}
\texttt{Vfin} \IMPL
	& \modM{1} (\texttt{single} \AND \texttt{filled}) \AND \modallM{1} \texttt{initial} & [1]\\
\AND	& \modallM{2} (\texttt{middle} \AND \texttt{norel}) & [2] \\
\AND	& \modM{3} \texttt{single} \AND \modallM{3} (\texttt{final} \AND \texttt{norel}) & [3] \\
\AND	& \texttt{V2} \EQUV (\texttt{middle} \AND	<_{*} \AND \modallM{1} \texttt{norel}) & [4] \\ 
\AND	& \texttt{VEnd} \EQUV (\texttt{middle} \AND >_{*}) & [5] \\
\AND	& \texttt{V1} \EQUV (\texttt{initial} \AND \texttt{norel}) & [6] \\
\end{array}
$
\caption{Domain Description of finite verbs
	\label{fig:Vfin}}
\end{figure}

The formula in Fig.\ref{fig:Vfin} states that all finite verbs (word class
\texttt{Vfin} $\in \C$) define three order domains, of which the first
requires exactly one element with the feature \texttt{initial} [1], the
second allows an unspecified number of elements with features
\texttt{middle} and \texttt{norel} [2], and the third allows at most one
element with features \texttt{final} and \texttt{norel} [3]. The features
\texttt{initial}, \texttt{middle}, and \texttt{final} $\in \A$ serve to
restrict placement of certain phrases in specific fields; e.g., no
reflexive pronouns can appear in the final field. The \texttt{norel} $\in
\A$ feature controls placement of a relative NP or PP, which may appear in
the initial field only in verb-final clauses. The order types are defined
as follows: In a verb-second clause (feature \texttt{V2}), the verb is
placed at the beginning ($<_{*}$) of the middle field (\texttt{middle}),
and the element of the initial field cannot be a relative phrase
($\modM{1}$\texttt{norel} in [4]). In a verb-final clause (\texttt{VEnd}),
the verb is placed at the end ($>_{*}$) of the middle field, with no
restrictions for the initial field (relative clauses and non-relative
verb-final clauses are subordinated to the noun and conjunction, resp.)
[5]. In a verb-initial clause (\texttt{V1}), the verb occupies the initial
field [6].

\begin{figure}[t]
\centering \small
$
\begin{array}{r@{}r@{\,}lc}
\mcl{3}{\mbox{\os{hat} } \AND \texttt{Vfin}} & [7]\\
	\AND	& \tuple{\texttt{subj}} (
			& \mbox{\os{Junge} } \AND \disko_\emptyset) & [8] \\
	\AND	& \tuple{\texttt{vpart}} (
			& \mbox{\os{gesehen} } \AND \disko_\emptyset & [9] \\
		& 	& \AND \ \neg\texttt{final}\  \AND \  >_{\{\texttt{subj}, \texttt{obj}\}}& [10] \\
		& 	& \AND \tuple{\texttt{obj}} (\mbox{\os{Mann} } \AND \disko_{\{\texttt{vpart}\}})) & [11]\\
\end{array}
$
\caption{Hierachical Structure
	\label{fig:hierarchy}}
\end{figure}

The formula in Fig.\ref{fig:hierarchy} encodes the hierarchical structure
from Fig.\ref{fig:dep-tree} and contains lexical restrictions on placement
and extraction (the surface is used to identify the word). Given this, the
order type of \os{hat} is determined as follows: The participle may not be
extraposed ($\neg\texttt{final}$ in [10]; a restriction from the lexical
entry of \os{hat}), it must follow \os{hat} in $d_2$. Thus, the verb cannot
be of order type \texttt{VEnd}, which would require it to be the last
element in its domain ($>_{*}$ in [5]). \os{Mann} is not adjacent to
\os{gesehen}, but may be extracted across the dependency \texttt{vpart}
($\disko_{\{\texttt{vpart}\}}$ in [11]), allowing its insertion into a
domain defined by \os{hat}. It cannot precede \os{hat} in $d_2$, because
\os{hat} must either begin $d_2$ (due to $<_{*}$ in [4]) or itself go into
$d_1$. But $d_1$ allows only one phrase (\texttt{single}), leaving only the
domain structure from Fig.\ref{fig:dom-tree}, and thus the order type
\texttt{V2} for \os{hat}.


\section{Comparison to PSG Approaches
	\label{ch:compare}}

One feature of word order domains is that they factor ordering alternatives
from the syntactic tree, much like feature annotations do for morphological
alternatives. Other lexicalized grammars collapse syntactic and ordering
information and are forced to represent ordering alternatives by lexical
ambiguity, most notable L-TAG \cite{Schabes1988} and some versions of CG
\cite{Hepple1994}. This is not necessary in our approach, which drastically
reduces the search space for parsing.

This property is shared by the proposal of \textcite{Reape-DISS} to
associate HPSG signs with sequences of constituents, also called word order
domains. Surface ordering is determined by the sequence of constituents
associated with the root node. The order domain of a mother node is the
sequence union of the order domains of the daughter nodes, which means that
the relative order of elements in an order domain is retained, but material
from several domains may be interleaved, resulting in discontinuities.
Whether an order domain allows interleaving with other domains is a
parameter of the constituent. This approach is very similar to ours in that
order domains separate word order from the syntactic tree, but there is one
important difference: Word order domains in HPSG do not completely free the
hierarchical structure from ordering considerations, because discontinuity
is specified per phrase, not per modifier. For example, two projections are
required for an NP, the lower one for the continuous material (determiner,
adjective, noun, genitival and prepositional attributes) and the higher one
for the possibly discontinuous relative clause. This dependence of
hierarchical structure on ordering is absent from our proposal.


We may also compare our approach with the projection architecture of LFG
\cite{Kaplan+Bresnan1982,Kaplan1995}. There is a close similarity of the
LFG projections (c-structure and f-structure) to the dimensions used here
(order domain structure and dependency tree, respectively). C-structure and
order domains represent surface ordering, whereas f-structure and
dependency tree show the subcategorization or valence requirements. What is
more, these projections or dimensions are linked in both accounts by an
element-wise mapping. The difference between the two architectures lies in
the linkage of the projections or dimensions: LFG maps f-structure off
c-structure. In contrast, the dependency relation is taken to be primitive
here, and ordering restrictions are taken to be indicators or consequences
of dependency relations (see also \cn{nobi:lfg98}
(\cy{nobi:lfg98,nobi:coling-dg98})).

\section{Conclusion
	\label{ch:conclu}}

We have presented an approach to word order for DG which combines
traditional notions (semantically motivated dependencies, topological
fields) with contemporary techniques (logical description language,
model-theoretic semantics). Word order domains are sets of partially
ordered words associated with words. A word is contained in an order domain
of its head, or may float into an order domain of a transitive head,
resulting in a discontinuous dependency tree while retaining a projective
order domain structure. Restrictions on the floating are expressed in a
lexicalized fashion in terms of dependency relations. An important benefit
is that the proposal is lexicalized without reverting to lexical ambiguity
to represent order variation, thus profiting even more from the efficiency
considerations discussed by \textcite{Schabes1988}.

It is not yet clear what the generative capacity of such lexicalized
discontinuous DGs is, but at least some index languages (such as
$a^nb^nc^n$) can be characterized. \textcite{nobi:acl97} have shown that
recognition and parsing of such grammars is \NP-complete. A parser
operating on the model structures is described in \cite{nobi:iwpt97}. 



\smallbibliography{\small}
\bibliography{ref,nobi,add-on}
\bibliographystyle{nobibib}

\end{document}